\documentclass[sigconf, anonymous=False]{acmart}
\usepackage[normalem]{ulem}
\useunder{\uline}{\ul}{}
\usepackage{algorithm}
\usepackage{algorithmic}
\usepackage{multirow}
\usepackage{amsmath} 
\usepackage{enumitem}
\usepackage{makecell}
\usepackage[most]{tcolorbox}

\usepackage{caption} % 载入caption宏包
\usepackage{etoolbox} % 载入etoolbox宏包

\AtBeginEnvironment{tabular}{\small}

\begin{document}
\newcommand{\qa}[1]{{\color{black}#1}}

\title{LeKUBE: A Legal Knowledge Update BEnchmark}

\author{Changyue Wang}
\email{changyuewang02@gmail.com}
\affiliation{
Department of Computer Science and Technology, Tsinghua University,
Beijing 100084 \country{China}
}

\affiliation{
Quancheng Lab\\
Jinan 250100 \country{China}
}

\author{Weihang Su}
\affiliation{
Department of Computer Science and Technology, Tsinghua University,
Beijing 100084 \country{China}
}

\author{Yiran Hu}
\affiliation{
School of Law, Tsinghua University,
Beijing 100084 \country{China}
}

\author{Qingyao Ai}
\email{aiqy@tsinghua.edu.cn}
\affiliation{
Department of Computer Science and Technology, Tsinghua University,
Beijing 100084 \country{China}
}

\author{Yueyue Wu}
\affiliation{
Department of Computer Science and Technology, Tsinghua University,
Beijing 100084 \country{China}
}

\author{Cheng Luo}
\authornote{Corresponding author}
\affiliation{
Quancheng Lab\\
Jinan 250100 \country{China}
}
\affiliation{
MegaTech.AI\\
Beijing 100080 \country{China}
}

\author{Yiqun Liu}
\affiliation{
Department of Computer Science and Technology, Tsinghua University,
Beijing 100084 \country{China}
}

\author{Min Zhang}
\affiliation{
Department of Computer Science and Technology, Tsinghua University,
Beijing 100084 \country{China}
}

\author{Shaoping Ma}
\affiliation{
Department of Computer Science and Technology, Tsinghua University,
Beijing 100084 \country{China}
}

\begin{abstract}

Recent advances in Large Language Models (LLMs) have significantly shaped the applications of AI in multiple fields, including the studies of legal intelligence. 
Trained on extensive legal texts, including statutes and legal documents, the legal LLMs can capture important legal knowledge/concepts effectively and provide important support for downstream legal applications such as legal consultancy. Yet, the dynamic nature of legal statutes and interpretations also poses new challenges to the use of LLMs in legal applications. Particularly, how to update the legal knowledge of LLMs effectively and efficiently has become an important research problem in practice.
Existing benchmarks for evaluating knowledge update methods are mostly designed for the open domain and cannot address the specific challenges of the legal domain, such as the nuanced application of new legal knowledge, the complexity and lengthiness of legal regulations, and the intricate nature of legal reasoning.
% To address this gap, we introduce the Legal Knowledge Update BEnchmark, i.e. LeKUBE, specifically tailored for the legal domain and focusing on Chinese law. 
% LeKUBE evaluates the effectiveness of different knowledge update methods. 
% It contains simulated updates of laws and legal knowledge and a series of tasks designed to assess whether the LLMs can accurately apply the new knowledge after knowledge updating. 
To address this gap, we introduce the Legal Knowledge Update BEnchmark, i.e. LeKUBE, which evaluates knowledge update methods for legal LLMs across five dimensions.\footnote{All the datasets and codes for evaluation are available at: https://github.com/bebr2/LeKUBE} 
Specifically, we categorize the needs of knowledge updates in the legal domain with the help of legal professionals, and then hire annotators from law schools to create synthetic updates to the Chinese Criminal and Civil Code as well as sets of questions of which the answers would change after the updates.  
Through a comprehensive evaluation of state-of-the-art knowledge update methods, we reveal a notable gap between existing knowledge update methods and the unique needs of the legal domain, emphasizing the need for further research and development of knowledge update mechanisms tailored for legal LLMs.

\end{abstract}


\maketitle

\section{Introduction}

In recent years, Large Language Models (LLMs) have demonstrated remarkable capabilities in a wide range of natural language processing (NLP) tasks. Particularly in the legal field, a considerable number of legal LLMs have been developed and used in legal practice today. Trained on a comprehensive legal corpus encompassing statutes, legal textbooks, and legal documents, these models possess an in-depth understanding of legal knowledge, which enables them to solve many legal problems in practice.
Yet, despite these advancements, legal LLMs face significant challenges due to the dynamic nature of legal statutes and administrative documents. This necessitates efficient and reliable methods for updating the internal knowledge of legal LLMs to reflect the latest changes in the legal field.

Knowledge updating has been widely considered to be an important research question for the construction of LLMs in open domains.
There have already been many datasets built focusing on benchmarking the performance of LLM knowledge editing methods in open domains. For example, CounterFact\cite{meng2022locating}, MQUAKE-CF\cite{zhong2023mquake}, and EVEDIT\cite{liu2024evedit} evaluate the ability of LLMs to revise their knowledge base when confronted with conflicting information. MQUAKE-T\cite{zhong2023mquake} and TimeSensitive-QA\cite{chen2021dataset} assess how LLMs handle and update obsolete information, probing their awareness of real-world temporal changes. Moreover, FreshQA\cite{vu2023freshllms} examines the capacity of LLMs to incorporate the most recent information effectively and precisely, with a focus on the timeliness of knowledge updates.

However, adapting general domain knowledge updating techniques to the legal domain introduces significant challenges due to the specific requirements of legal knowledge application. 
Firstly, legal knowledge updates, particularly those on legal statutes, often have higher requirements on update effectiveness and precisions than those in open domains.
Secondly, legal texts are often intricate and complex, which presents difficulties for LLMs in tasks that involve paraphrasing extensive legal documents. 
Furthermore, in contrast to knowledge editing in open domains, updates in legal knowledge usually have profound impacts on legal reasoning, thereby influencing the outcomes of numerous legal tasks.
These considerations highlight the limitations of current knowledge update benchmarks from general domains when applied to legal LLMs, which fail to capture the unique challenges and specifics of the legal field.

To address this gap, we introduce the Legal Knowledge Update Benchmark, i.e. LeKUBE, which is specifically tailored for the legal domain and focuses on Chinese law. 
With the help of legal professionals, LeKUBE categorizes the needs of knowledge updates in the legal domains into five dimensions and develops evaluation tasks for each of them separately.
Specifically, we hired high-quality annotators from law schools to create synthetic updates to Chinese Criminal Law and Civil Code and corresponding questions and tasks that evolve these updates. These data not only support the evaluation of knowledge updating methods in legal LLMs, but also provide important insights on understanding the needs of legal knowledge updates in practical legal applications.

We conduct experiments on a wide range of state-of-the-art knowledge update baselines on LeKUBE. 
The selected baselines can be divided into non-parametric strategies and parametric strategies. 
Non-parametric strategies do not alter the model’s parameters but inject new knowledge into the input of the LLM. 
In contrast, parametric strategies inject knowledge by changing the parameters of the model.
Our experimental results not only reveal a varied performance of different knowledge update methods across tasks but also showcase the differences and difficulty of legal knowledge updating. 
% Overall, non-parametric strategies outperform parametric ones, but they also have drawbacks such as increased inference time, slightly poor generality, and knowledge conflicts. Among parametric strategies, each method has its strengths and weaknesses. For instance, Lora fine-tuning\cite{hu2021lora} far outperforms other parametric strategies in terms of time efficiency, while the Self-Edit\cite{liu2024evedit} method excels in enhancing reasoning ability with new knowledge.

% To address the challenges of knowledge updates in legal LLMs, we introduce a new legal knowledge update benchmark, LeKUBE (Legal Knowledge Update BEnchmark). The LeKUBE benchmark is specifically designed to assess the effectiveness of knowledge update methods in the legal domain, covering multiple tasks and capability dimensions. We then compare the performance of different knowledge update methods on legal LLMs through experiments and provide a detailed analysis.

% To address the challenges of knowledge updates in legal language models (LLMs), we introduce the Legal Knowledge Update BEnchmark (LeKUBE). Specifically designed for Chinese law, LeKUBE assesses the efficacy of knowledge update methods through a series of tasks that simulate legal updates and test how well these methods enable LLMs to integrate new knowledge. 
% By evaluating whether LLMs have effectively acquired new legal knowledge, the LeKUBE benchmark provides detailed analysis and comparison of different knowledge update techniques across multiple tasks and capability dimensions.

In summary, the main contributions of this paper are as follows:
\begin{itemize}[leftmargin=*]
\item We reveal the potential difficulties and challenges of knowledge updates in legal LLMs and point out that existing general domain benchmarks cannot adequately evaluate the performance of knowledge update methods in the legal field.

\item We propose the LeKUBE benchmark, a knowledge update benchmark tailored for evaluating knowledge update methods in the legal domain.\footnote{https://github.com/bebr2/LeKUBE}

\item We systematically evaluate the performance of various existing knowledge update methods in the legal domain and provide a detailed comparative analysis, offering reference and guidance for the future optimization and development of knowledge update technology in the legal domain.

\end{itemize}

\section{Related Work}

% \subsection{Legal LLMs}

% In the legal field, the development of legal-specific Large Language Models (LLMs) has been propelled by the demand for specialized knowledge. \textbf{ChatLaw} \cite{cui2023chatlaw}, \textbf{Lawyer LLaMA} \cite{huang2023lawyer}, and \textbf{LegalAID} \cite{megatechai2023legalaid} are key examples of such models. ChatLaw, an open-source legal LLM, combines vector database retrieval with keyword search to extract legal knowledge and introduces a "self-suggestion" mechanism to reduce the model's hallucination. Lawyer LLaMA infuses legal domain knowledge during the continuous training phase, uses a Supervised Fine-Tuning (SFT) dataset for instruction following, and incorporates a retrieval module to extract relevant statutes. LegalAID imitates the learning process of law students, introducing a curriculum-based learning approach that involves various stages of legal study, from memorizing regulations to reasoning legal inference.

% Despite these advancements, the application of LLMs in the legal field has revealed certain deficiencies. The training of these models requires substantial resources and time, and when legal knowledge undergoes updates, the models need to be updated accordingly. The cost-effectiveness of knowledge updates in the field of legal LLMs has not been adequately studied.

\subsection{Knowledge Update Methods for LLMs}

% In the context of LLMs, the need for efficient knowledge update mechanisms is important, given the rapid evolution of real-world data and the consequent obsolescence of certain knowledge within models. Broadly, these mechanisms fall into two categories: parametric and non-parametric update strategies.

% \subsubsection{Parametric Update Strategies}

Existing knowledge update methods can be divided into two categories, parametric and non-parametric.
Parametric strategies involve the direct modification of model parameters to update knowledge. Two principal techniques exist within this category: model fine-tuning and model editing.
Fine-tuning is a process where the model parameters are adjusted based on a specific task or dataset. 
The objective is to minimize the loss between the model's predictions and the fine-tuning dataset. Fine-tuning can be further classified into full-parameter fine-tuning, which adjusts all parameters of the model, and Lora fine-tuning \cite{hu2021lora}, a more resource-efficient method that introduces a low-rank structure in the model's weight matrix for adjustment.
On the other hand, model editing techniques aim to precisely modify the model parameters that influence the integration of new knowledge. This method effectively incorporates new knowledge while preserving unrelated knowledge within the model. Techniques in this category include Knowledge Neurons (KN) \cite{dai2021knowledge}, which identifies and edits "knowledge neurons" in the pre-trained model, Rank-One Model Editing (ROME) \cite{meng2022locating}, which uses causal tracing to directly write new key-value pairs into the earlier feed-forward network (FFN) layers of the model, and Self-Edit \cite{liu2024evedit}, which defines editing boundaries using event context and improves editing consistency while maintaining the naturalness of the generated text.

Non-parametric strategies allow for flexible knowledge updates without the need for retraining the entire model. A notable example is Retrieval-Augmented Generation (RAG) \cite{su2024dragin,lewis2020retrieval,gao2023retrieval}, which integrates results retrieved from a knowledge base to assist the generation process of the language model. 
RAG has been demonstrated to enhance the performance of LLMs and alleviate hallucinations \cite{shuster2021retrieval,su2024unsupervised,su2024mitigating}.
With this approach, knowledge updates only require modifications to the knowledge base, not the internal parameters of the model.
While these strategies offer promising results, they also present challenges, such as handling noise in retrieval results \cite{cuconasu2024power}, managing interactions between the retriever and generator \cite{zhao2024retrieval}, and addressing context limitations in long text generation \cite{jiang2023llmlingua}.

\subsection{Evaluation of Knowledge Updating}

Analyzing the efficiency of knowledge update methods for LLMs is crucial. As Wang et al.\cite{wang2023knowledge} highlighted, these methods can be evaluated across five dimensions: accuracy, locality, generality, Retainability, and Scalability.
\textbf{Accuracy}\cite{dong2022calibrating, meng2022locating, dai2021knowledge, zhong2023mquake} gauges the ratio of successful updates in the dataset. Dong et al.\cite{dong2022calibrating} introduced the CKA framework to assess accuracy by comparing the scores of correct and incorrect facts.
\textbf{Locality}\cite{mitchell2022memorybased, sinitsin2020editable, zhong2023mquake} assess how a method retains unrelated knowledge during specific updates. The Drawdown metric by Sinitsin et al. \cite{sinitsin2020editable} measures locality. And SERAC\cite{mitchell2022memorybased} further refined this evaluation method using the similarity of embedding vectors.
\textbf{Generality}\cite{liu2024evedit, meng2022locating, zhong2023mquake} tests if updated knowledge can be generalized to other relevant inputs. EVEDIT\cite{liu2024evedit} and MQUAKE\cite{zhong2023mquake} test generality through inferential questions and multi-hop questions, respectively.
{Retainability}\cite{huang2023transformerpatcher} aims to check if a method can maintain the effects of early updates after multiple updates. Transformer-Patcher\cite{huang2023transformerpatcher} explored this aspect.
{Scalability}\cite{yao2023editing} evaluates if a method can handle large-scale updates. Yao et al.\cite{yao2023editing} emphasized the importance of scalability, especially for real-time updating and maintaining large knowledge bases.

\subsection{Existing Benchmarks}
% \subsection{Benchmark for Evaluation of Knowledge Update Methods in General Domain}

A variety of datasets have been developed to thoroughly evaluate the performance of knowledge update techniques in general domains. These datasets span different fields and task types and can be categorized into three main groups based on their construction methods and the types of knowledge updates they focus on evaluating counterfactual update capacity, assessing time awareness, and examining the ability to acquire the latest information.
For evaluating counterfactual update capacity, datasets such as CounterFact\cite{meng2022locating}, MQUAKE-CF\cite{zhong2023mquake}, and EVEDIT\cite{liu2024evedit} have been developed. These datasets focus on testing the ability of LLMs to handle information that contradicts their prior knowledge, thereby enhancing the robustness and adaptability of knowledge update techniques.
Datasets like MQUAKE-T\cite{zhong2023mquake} and TimeSensitive-QA\cite{chen2021dataset} are designed to assess temporal evolution awareness. They focus on how well LLMs can understand and update outdated information, reflecting the model's awareness of the temporal evolution of real-world information.
Finally, for examining the ability to acquire the latest information, datasets like FreshQA\cite{vu2023freshllms} have been created. These datasets focus on how effectively knowledge update techniques can inject the latest knowledge into LLMs in a timely and accurate manner.

% These benchmarks provide a comprehensive evaluation of the effectiveness of various knowledge update methods in general domains, contributing to the ongoing development of more robust and adaptable LLMs.

% \section{Methodology}

% \begin{figure}[t!]
% \includegraphics[width=0.5\textwidth]{pic/compare.pdf}
% \caption{In the example of CounterFact, answering the question requires reasoning through two knowledge triples: (Turkish Historical Society, located in, Gazi University) and (Gazi University, located in, Glasgow). In the example of LeKUBE, it is necessary to extract judgment elements from the case, such as the Walkman and Li, and then abstract them into legal concepts, such as lost property and rights holder, etc. Based on this, the LLM will look for the corresponding statutes in legal documents, such as Article 312 of the Civil Code, and finally summarize the case and the legal provisions to make a judgment.}
% \label{fig:compare_rome_lekube}
% \end{figure}

\section{Challenges of Legal Knowledge Update}

\label{cha:challenge}

Updating knowledge in the legal domain presents unique challenges that are not fully addressed by current benchmarks for the general one. These challenges arise from the specific characteristics of legal knowledge and the way it is applied and updated. 

\begin{itemize}[leftmargin=*]
\item \textbf{Application of New Legal Knowledge}: One primary challenge in the legal domain is the correct application of new legal knowledge. Unlike general LLMs that respond to temporal changes by emphasizing the application of the latest information, the legal domain requires careful consideration of specific case circumstances. Factors such as the time span, differences between new and old laws, and the severity of penalties must be taken into account to determine whether to apply new or old statutes. This requirement significantly complicates the task of knowledge updating.
Furthermore, beyond merely updating knowledge, legal LLMs should also learn the patterns and evolution of legal changes. This deeper understanding is essential for accurately navigating the complexities of legal reasoning and judgments over time.
    
\item  \textbf{Complexity of Legal Regulations}: Legal regulations are typically more detailed and complex than the entity names in the general domain. Existing benchmarks for evaluating the accuracy of knowledge update often require models to paraphrase short entity names. However, when the legal domain requires the LLM to paraphrase lengthy legal regulations, it poses a significant challenge to existing knowledge update methods. For example, the average length of the new targets in the CounterFact dataset is 6.66 characters in English, while the average length of the articles in the Criminal Law of the People's Republic of China is 135.63 characters in Chinese.

\item  \textbf{Impact on Legal Reasoning}: Updates to legal knowledge can influence legal reasoning, consequently affecting the results of many legal tasks. Although some general domain benchmark datasets, such as CounterFact, involve reasoning questions after knowledge updating, they remain at the level of entity relationship reasoning. This does not satisfy the requirements for legal reasoning, which often involves extraction and abstraction of case elements, matches of legal provisions, etc. 
% \autoref{fig:compare_rome_lekube} shows the difference in reasoning capabilities examined by CounterFact and LeKUBE.

\end{itemize}

In summary, the unique characteristics of the legal domain, such as the application of new laws, the complexity of legal regulations, and the profound impact of legal knowledge updates on legal reasoning, present significant challenges to the current knowledge update methods. These challenges highlight the need for more sophisticated models and evaluation benchmarks tailored specifically to the legal domain.

\section{Evaluation Dimensions of LeKUBE}

% In this section, we introduce the legal knowledge update tasks in LeKUBE. These tasks are designed to evaluate the effectiveness of knowledge update methods in the legal LLM, modeled after those in the general domain. 

This section introduces the evaluation dimensions in LeKUBE, designed to assess the effectiveness of knowledge update methods in legal LLMs, modeled after those in the general domain\cite{wang2023knowledge}.

\subsection{Accuracy}

In the general domain of knowledge update, accuracy measures the proportion of data that is successfully updated, i.e., the proportion of the updated entity that the LLM can successfully paraphrase when given the prompt. In the legal domain, the challenge for updating the LLM is the complete memorization of the updated legal knowledge. In LeKUBE, the main updated knowledge is the legal statutes, and this part involves two tasks:
\begin{itemize}[leftmargin=*]
  \item \textbf{Recitation of Statutes}: Given the updated statute, the LLM needs to accurately recite the specific content of the statute.
  \item \textbf{Recall of Statute Items}: The dual task of recitation of statutes. Given the specific content of the updated statute, the LLM needs to answer which clause of which law the statute comes from.
\end{itemize}

% As mentioned above, the legal domain has much higher requirements for accuracy than the general domain. The evaluation benchmark for knowledge update in the general domain usually only requires the LLM to correctly answer the name of the updated entity, and the name of the entity can have different expressions. However, in the legal domain, the content of statutes is usually much longer than the name of the entity, and its expression is unique, which poses a great challenge to the knowledge update method.

\subsection{Generality}

Generality requires that the updated LLM not only perform well on the training dataset but also generalize the relevant knowledge to other inputs. In the general domain, the examinations generally fall into two dimensions: examining the LLM's sense of time, or examining the LLM's inference of updated knowledge. In LeKUBE, we require the LLM to complete the following two tasks:
\begin{itemize}[leftmargin=*]
  \item \textbf{True-or-False Questions of Change in Statute}: For the updated legal statute, the LLM needs to judge whether a certain detail of the statute has changed compared to before. This task requires the LLM to have a good grasp of the revision process and details of legal texts. This is the basis for the LLM to understand the application of the new statutes. 
  % In the general domain, there are also datasets specifically for evaluating the time consciousness of the updated LLM, such as xxxx, but they focus on examining the LLM's understanding of time, rather than comparing the differences in the details of old and new knowledge.
  \item \textbf{Multiple-Choice Questions (MCQ) of the Legal Scenario}: For each updated statute, we designed a choice question that incorporates legal scenarios or virtual cases. The answer to this question changes before and after the statute is updated. We hope that the updated LLMs can correctly answer the updated answer. This task examines the practical application and scenario inference of the updated legal content.
\end{itemize}

\subsection{Locality}
\label{cha:local}

Locality requires that the updated LLM still maintain a good grasp of other unchanged knowledge, that is, the knowledge update method should not only perform well on tasks related to updated knowledge but also not affect the LLM's original performance of unrelated knowledge. The legal domain requires the updated LLM to maintain the memory of the unchanged statutes, so in this part, the tasks we designed are the Recitation of Statutes, the Recall of Statute Items, and the True-or-False Questions of Change in Statute, related to statutes that have not been updated.

% \begin{figure*}[t!]
% \begin{minipage}[t]{0.45\textwidth}
% \hspace{-1.5cm}
% \includegraphics[width=10cm]{pic/scale_.pdf}
% \caption{Flowchart illustrating our scalability evaluation by comparing two methodologies. In the blue section, individual LLMs are updated based on knowledge from specific chapters, and their performance is evaluated by average question accuracy. In the red section, a single LLM is updated with knowledge from all chapters that need to be updated in LeKUBE.}
% \label{fig:intro_scale}
% \end{minipage}
% \begin{minipage}[t]
% {0.45\textwidth}
% \hspace{-0.5cm}
% \includegraphics[width=10cm]{pic/retain_.pdf}
% \caption{Flowchart illustrating our retainability evaluation by comparing two phases. The first one is the situation where the criminal law is updated while the civil code is not. And the second one is the situation where the two laws are both updated.}
% \label{fig:intro_retain}
% \end{minipage}

% \end{figure*}

\begin{figure}[t!]
\includegraphics[width=1\columnwidth]{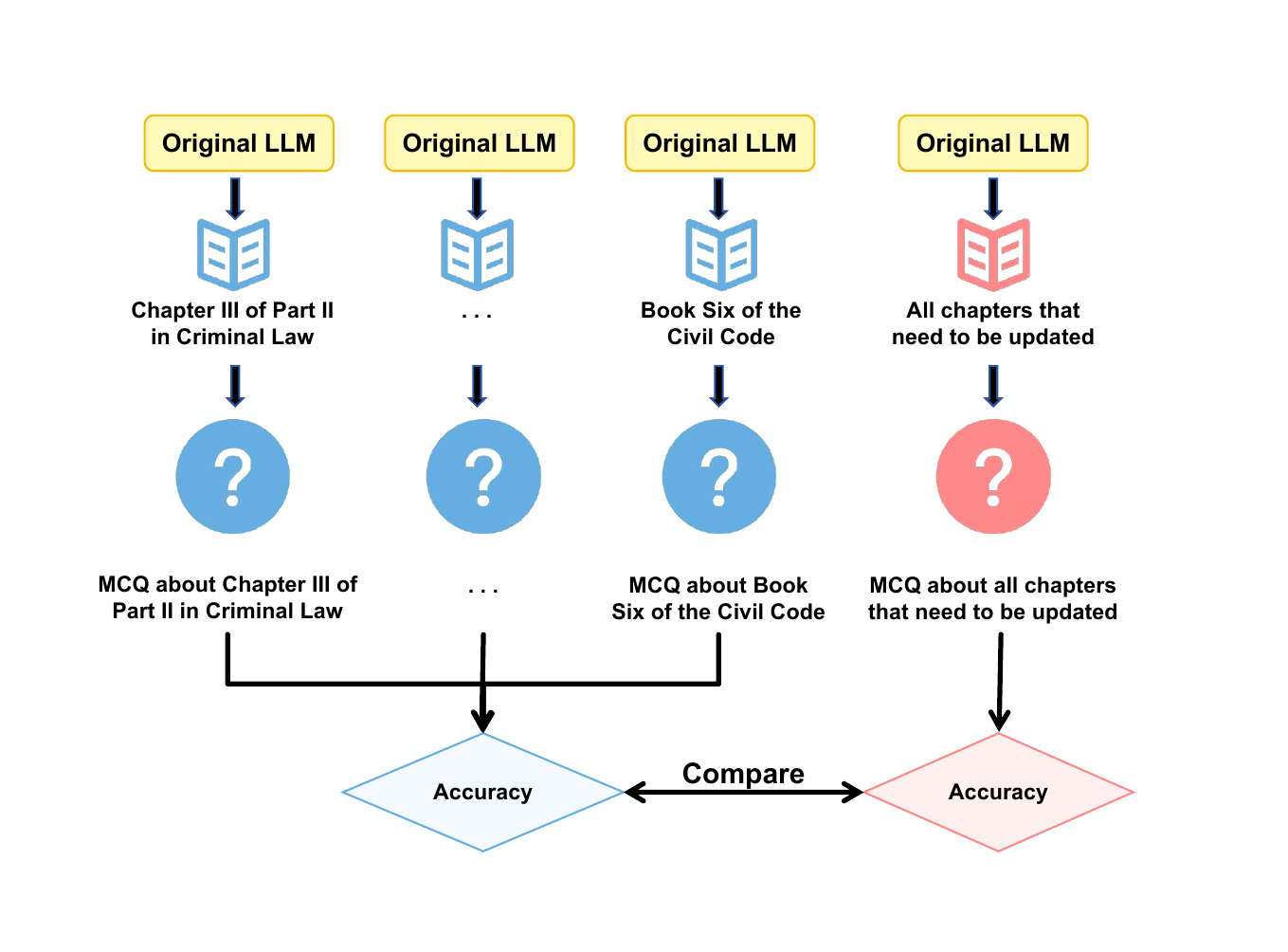}
\caption{Flowchart illustrating our scalability evaluation by comparing two methodologies. In the blue section, individual LLMs are updated based on knowledge from specific chapters, and their performance is evaluated by average question accuracy. In the red section, a single LLM is updated with knowledge from all chapters that need to be updated in LeKUBE.}
\label{fig:intro_scale}
\end{figure}

\subsection{Scalability}

Scalability evaluates the relationship between the effect of knowledge update and the amount of updated knowledge. If, When the amount of updated knowledge increases, the effect of knowledge update does not significantly decrease, then we believe that the knowledge update method has good scalability. In LeKUBE, we divide the data into chapters (such as "Chapter III of Part II in Criminal Law", "Book Six of the Civil Code", etc.). For a certain knowledge update method, we let the LLM update the knowledge by chapters (i.e., assuming that the statutes are only updated within a single chapter), and then complete the Multiple-Choice Questions of the Legal Scenario of this chapter. We compare this part of the performance with the performance of the LLM updating all the chapters that need to be updated, and the difference in performance reflects the scalability of the knowledge update method. \autoref{fig:intro_scale} show the process when evaluating scalability.

\subsection{Retainability}

\begin{figure}[t!]
\includegraphics[width=1\columnwidth]{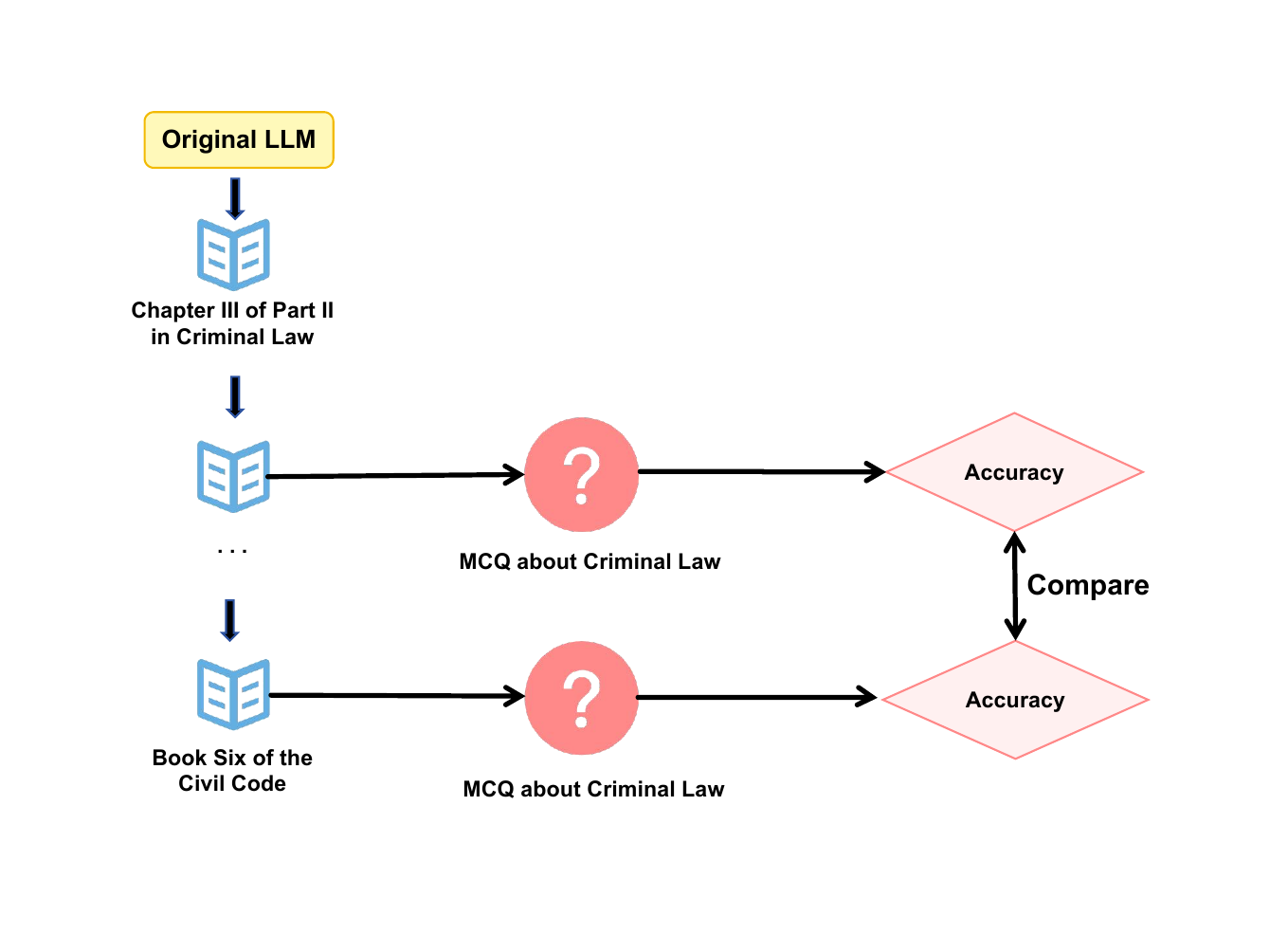}
\caption{Flowchart illustrating our retainability evaluation by comparing two phases. The first one is the situation where the criminal law is updated while the civil code is not. And the second one is the situation where the two laws are both updated.}
\label{fig:intro_retain}
\end{figure}

Retainability evaluates the degree to which the LLM retains the effect of early updates after multiple updates. Similarly, we use chapters as units and update the knowledge in a predetermined order. Then we choose the task of Multiple-Choice Questions of the Legal Scenario and evaluate the performance of the LLM after updating some data and updating all the data. The difference in performance reflects the retainability of the knowledge update method. \autoref{fig:intro_retain} show the process when evaluating retainability.

\section{Dataset Construction}

\subsection{Data Source}

% We focus on Chinese laws in this paper. The Chinese legal system includes two important statutes, namely, the Criminal Law of the People's Republic of China (hereinafter referred to as the Criminal Law) and the Civil Code of the People's Republic of China (hereinafter referred to as the Civil Code). The statutes for modification in the dataset are selected from the latest versions of these two laws. The raw data is from STARD\cite{su2024stardchinesestatuteretrieval}, which collected all national-level laws and regulations and judicial interpretations in China from official government sources.
In this paper, we focus on the Chinese legal system, specifically targeting two principal statutes: the Criminal Law and the Civil Code of the People’s Republic of China, referred to as the Criminal Law and the Civil Code, respectively. 
The statutes incorporated into our dataset are derived from the most recent versions of these laws. Our primary data source is the corpus of the STARD dataset \cite{su2024stard}, which encompasses a comprehensive collection of all national-level laws, regulations, and judicial interpretations in China, sourced directly from official government platforms.

% Our annotation team first listed the Criminal Law and the Civil Code in China, then manually downloaded the latest versions from official government sources.
\subsection{Annotation Recruitment and Payment}

We recruit expert annotators from top law schools for dataset annotation tasks. Annotators are paid based on the number of complete annotations, i.e., pairs of annotations involving a statute modification and a corresponding legal multiple-choice question. The average payment per annotation is 10 CNY. Typically, an annotator can complete 4-5 annotations per hour, resulting in an average hourly wage of 40-50 CNY, which is over 80\% higher than Beijing's minimum wage. We also employ other annotators for quality evaluation of the annotations and questions generated by the LLMs, at an average payment of 3 CNY per evaluation. An evaluator can typically assess 15 annotations per hour, resulting in an average wage of 45 CNY per hour.

\subsection{Annotation Process}

\subsubsection{Statute Modification and Generation of Multiple-Choice Questions}

To ensure diverse and non-conflicting legal modifications, chapters from Criminal Law and Civil Code are assigned to different annotators who are tasked to modify parts of the statutes. They then proposed a multiple-choice question related to the modification. 

For simplicity, we only allow three types of modifications in our annotation process, i.e., changing legal consequences, changing constituent elements, and changing behavior patterns. Annotators are required to select one method before modifying a statute and provide a brief reason for their modification. Also, we require that the modified statutes should be reasonable, legally accurate, and diverse. The modification of related statutes should be consistent to avoid contradictions. And the multiple-choice questions should involve a legal scenario or a fictional case with an answer that changes based on the statute modification. Each data point is added to the final LeKUBE dataset only after another annotator has checked and confirmed their quality.

\subsubsection{Generation of True-or-False Questions of Change in Statute}

True-or-false questions are generated for each modified statute to assess the details of the statute change. Unmodified statutes are also included to get questions for evaluation of locality. 

We concatenate both the pre- and post-modification versions of each legal statute, including the modification method, into a single prompt for GPT-4-0125-preview\cite{openai2024gpt4}. Then we instruct the model to analyze the differences between these two versions and subsequently generate a series of true-or-false questions. Human annotators then filter the data, retaining only clear, reasonable, and correct questions and answers. The data is added to the final LeKUBE dataset only after another annotator approves of its quality.

% We use the GPT-4-0125-preview model\cite{openai2024gpt4} to generate the questions by providing it with two versions of a statute and the modification method. Human annotators then filter the data, retaining only clear, reasonable, and correct questions and answers. The data is added to the final LeKUBE dataset only after another annotator approves of its quality.

% \subsection{Annotation Consistency Analysis}

% To evaluate the reliability of consistency between annotators, we use Cohen's Kappa coefficient $ \mathcal{K} $ \cite{Cohen1960ACO} in a binary classification scenario. 
% Specifically, we randomly select 40 statute modifications and multiple-choice questions data from an annotator, and invite two other annotators to evaluate whether this data could be included in the final dataset, obtaining a $ \mathcal{K} $ value of 0.6078. We randomly select 200 LLM-generated true-or-false questions and let two annotators classify them for quality evaluation, obtaining a $ \mathcal{K} $ value of 0.7567. These indicate the effectiveness and consistency of our annotation process.

% \subsection{Ethical Discussions}

% LeKUBE serves only as a benchmark for evaluating knowledge update methods in the field of legal LLMs. The modifications to the statutes have no real-world implications, and the modified statutes and their multiple-choice questions do not affect any real-world legal practice.

\section{Dataset Statistics and Analysis}

Table \ref{tab:stat} and Table \ref{tab:stat2} provide a basic statistical overview of the LeKUBE dataset. LeKUBE contains 180 updated statutes, with the average length of each statute (calculated as Chinese characters) approximately 114 pre-update and around 124 post-update. Notably, some statutes in the dataset have been shortened after the update, such as those with certain clauses removed. Additionally, 60 non-updated statutes are randomly selected to evaluate the locality. The average length of these statutes is approximately 119, and other statistical indicators are presented in Table \ref{tab:stat2}.

\begin{table}[t]
  \centering
  \tiny
  \begin{tabular}{lll}
  \toprule
  \multicolumn{1}{l}{}              & \textbf{Statistics} & \textbf{\# Number} \\ \midrule
  \multirow{3}{*}{\textbf{Statutes}}      & Number of Statutes          & 180         \\
                                    & Average Length (before updated)          & 114.28     \\ 
                                    & Average Length (after updated)          & 124.36     \\ \hline
  \multirow{3}{*}{\makecell[c]{\textbf{True-or-False} \\  \textbf{Questions}}} & Number of Questions         & 642        \\
                                    & Average Questions per Statute  & 3.57       \\
                                    & Average Question Length        & 41.96      \\ \hline
    \multicolumn{1}{l}{\makecell[c]{\textbf{Multiple-Choice} \\  \textbf{Questions}}}              & Average Question Length & 49.12 \\
  \bottomrule
  \end{tabular}
  \caption{Statistics of Updated Statutes Data in LeKUBE}
  \label{tab:stat}
  \vspace{-4mm}
  \end{table}

\begin{table}[t]
  \centering
  \tiny
  \begin{tabular}{lll}
  \toprule
  \multicolumn{1}{l}{}              & \textbf{Statistics} & \textbf{\# Number} \\ \hline
  \multirow{2}{*}{\textbf{Statutes}}      & Number of Statutes          & 60         \\
                                    & Average Length          & 118.67     \\ \hline
  \multirow{3}{*}{\makecell[c]{\textbf{True-or-False} \\  \textbf{Questions}}} & Number of Questions         & 164        \\
                                    & Average Questions per Statute  & 2.73       \\
                                    & Average Question Length        & 43.38      \\
  \bottomrule
  \end{tabular}
  \caption{Statistics of Non-Updated Statutes Data in LeKUBE}
  \label{tab:stat2}
  % \vspace{-4mm}
  \end{table}

For the true-or-false questions of change in statute, LeKUBE contains 642 questions related to updated statutes and 164 questions related to non-updated statutes, the latter of which are designed to test the locality, as mentioned in \autoref{cha:local}. The former corresponds to an average of approximately 3.6 questions per statute, while the latter corresponds to around 2.7 questions per statute. The average length of the questions is about 45. Regarding the multiple-choice questions of the legal scenario, each updated statute corresponds to one question, with the average question length being approximately 42, slightly shorter than the true-or-false question length. Furthermore, we count the distribution of statute modification methods chosen by annotators. The counts for "changing constituent elements" and "changing legal consequences" are 26.7\% and 27.8\%, respectively, while the counts for "changing behavior patterns" are 45.6\%. This demonstrates that the methods of statute modification in the dataset are diverse. After rigorous quality filtering, LeKUBE comprises nine chapters in total, encompassing four chapters from Criminal Law (Chapters III, VI, VIII, IX) and five chapters from the Civil Code (Book 2-3, 5-7).

% \begin{figure}[htbp]
%     \centering
%     \includegraphics[width=0.5\textwidth]{pic/stat_1.png}
%       \caption{Distribution of Statute Modification Methods}
%       \label{fig:stat1}
% \end{figure} 

% Figure \ref{fig:stat3} shows the number of modified statutes in each chapter. Among the 9 involved chapters, the Civil Code shows significant variation in the number of modifications per chapter, while the Criminal Law exhibits a more uniform distribution. This variation correlates with the complexity of each chapter. It also demonstrates the diversity of the content of the statutes involved in LeKUBE.

% \begin{figure}[h!]
%   \centering
%   \includegraphics[width=0.45\textwidth]{pic/stat3.pdf}
%     \caption{Distribution of Updated Statutes by Chapter in LeKUBE(Horizontal Axis: Number; Vertical axis: Chapters that contain updated statutes in LeKUBE)}
%     \label{fig:stat3}
% \end{figure}

\section{Experimental Setup}

\subsection{Evaluation Procedure and Tasks}

We now provide a detailed description of the evaluation of different knowledge update methods on LeKUBE. Let the set of all statutes that need to be updated in LeKUBE be $\mathcal{D}_1$, and the set of statutes that haven't been updated in LeKUBE be $\mathcal{D}_2$.

In the three types of evaluation, which evaluate the accuracy, generality, and locality respectively, the set of statutes to update is $\mathcal{D}_1$. Then tasks for testing accuracy and generality are related to $\mathcal{D}_1$, and tasks for testing 
 locality are related to $\mathcal{D}_2$.

Then, for scalability and retainability, we divide $\mathcal{D}_1$ into $\mathcal{D}_1 = \mathcal{D}_1^1 \cup \mathcal{D}_1^2 \cup ...  \cup \mathcal{D}_1^9$ by chapters. The order is as follows: Chapter III, VI, VIII , and IX of Part II in the Criminal Law, and Book 2, 3, 5, 6, and 7 in the Civil Code. For tasks corresponding to scalability, we update the LLM on $\mathcal{D}_1^i, i = 1, ..., 9$ respectively, and evaluate the performance of the updated model $M_i$ on tasks corresponding to $\mathcal{D}_1^i$, then comparing it with the LLM updating the whole $D_1$ with the corresponding method. For tasks related to retainability, we sequentially update the model on $\mathcal{D}_1^i, i = 1, ..., 9$, and evaluate the performance difference of the model on tasks corresponding to $\mathcal{D}_1^i, i = 1,2,3,4$ after updating to $i=4$ (since there are four chapters of criminal law to update in LeKUBE, this corresponds to the situation where the criminal law is updated while the civil code is not).

% \subsection{Evaluation Metrics}

% \textbf{Exact Match (EM)}: For the tasks of Recitation of Statutes and Recall of Statute Items, we adopt Exact Match as the metric. Specifically, we first extract the answer part from the LLM's output, then retain only the Chinese text of the model's output, and perform an exact match with the ground truth, calculating the ratio of exactly matched answers. Particularly for Recall of Statute Items, as the LLM's response may be Arabic numerals, we allow Arabic numerals as answers when performing an exact match.

% \textbf{Accuracy (Acc)}: For Multiple-Choice and True-or-False questions, we use Accuracy as the metric. Specifically, we instruct the model to directly output the answer (such as the letter option for multiple-choice questions or "correct" or "incorrect" for true-or-false questions), then extract the answer from the LLM's output, and calculate the accuracy.

Depending on the task formats, the evaluation metrics we used include:

\textbf{Exact Match (EM)}: Employed for  Recitation of Statutes and Recall of Statute Items tasks. It involves extracting the Chinese text from the LLM's output and comparing it to the ground truth. The ratio of exact matches is computed. Arabic numerals are allowed in the Recall of Statute Items task.

\textbf{Accuracy (Acc)}: Employed for Multiple-Choice and True-or-False questions. The model is directed to output the answer, which is then extracted, and the accuracy is calculated based on the ground truth.

\subsection{Large Language Models}

% We use a total of five LLMs. The close-source model is the GLM4\cite{du2022glm} model, for which we use the official API for inference. The open-source models are BaiChuan2-13B-

We choose BaiChuan2-13B-Chat\cite{yang2023baichuan2openlargescale}, ChatGLM3-6B\cite{du2022glm}, ChatLaw-13B\cite{huang2023lawyer}, and LegalAID-7B\cite{su2023legalaid} for our experiments. The first two models are open-domain LLMs, while the latter two are Chinese legal LLMs.

% To ensure that knowledge updating is performed when LLMs already have old legal knowledge, we fine-tune the two general LLMs, BaiChuan2-13B-Chat and ChatGLM3-6B, on statutes and judicial interpretations to ensure that the models have sufficient legal knowledge.  
To ensure that knowledge updating is performed when LLMs already possess outdated legal knowledge, we further train BaiChuan2-13B-Chat and ChatGLM3-6B on the STARD dataset’s corpus, which encompasses a comprehensive collection of all national-level laws, regulations, and judicial interpretations in China. 
% his approach ensures that the models are equipped with up-to-date and sufficient legal knowledge.

% we ask professional annotators to list all national-level laws and regulations and judicial interpretations in China, and download the latest versions from official government sources, resulting in a total of 55,375 data points. We then fine-tune the models on these data, training for 2 epochs.

\subsection{Knowledge Update Baselines}

In this paper, the knowledge update baselines evaluated are categorized into two main types based on whether they alter or introduce new model parameters.

\subsubsection{Non-parametric Update Strategies}

The most popular non-parametric update strategy for LLMs is Retrieval Augmented Generation (RAG), which concatenates the retrieved legal text directly with the question as a prompt, injecting knowledge into the model in the form of context. 
Existing mainstream retrieval methods include vocabulary-based lexical-matching approaches ~\cite{ramos2003using, robertson2009probabilistic, zhai2008statistical} and dense retrieval~\cite{gao2021condenser, su2023wikiformer,fang2024scaling,ye2024relevance,li2023towards,chen2023thuir,su2023thuir2,chen2022web}. 
Due to the lack of domain-specific knowledge, dense retrieval models in open domains do not yield optimal results in legal search tasks. Consequently, considerable research has focused on better adapting dense retrieval methodologies to the legal domain~\cite{xiao2021lawformer,su2023caseformer,ma2023caseencoder,li2023thuir,li2023thuir2}.
In this paper, we selected both vocabulary-based lexical-matching method and dense retrieval method as our retrieval method:
% Depending on the type of retriever used, we choose two methods:

\begin{itemize}[leftmargin=*]
\item \textbf{RAG-BM25}: BM25 \cite{INR-019} is used as the retriever. BM25 is an efficient retrieval model based on lexical matching. In our evaluation, we employ an improved algorithm that adds two parameters to the base TF-IDF: term frequency saturation and field length normalization, with values set at 1.5 and 0.75 respectively.
\item \textbf{RAG-Lawformer}: Lawformer \cite{xiao2021lawformer} is used as the retriever. Lawformer, pre-trained based on Longformer \cite{beltagy2020longformer}, is specifically designed for understanding long Chinese legal documents. We use Lawformer to generate vector representations of queries and documents, and build an index of dense vectors. During retrieval, cosine distance is used to measure the similarity between queries and documents.
\end{itemize}

% \textbf{RAG-BM25}: BM25 \cite{INR-019} is used as the retriever. BM25 is an efficient retrieval model based on lexical matching. In our evaluation, we employ an improved algorithm that adds two parameters to the base TF-IDF: term frequency saturation and field length normalization, with values set at 1.5 and 0.75 respectively.

% \textbf{RAG-Lawformer}: Lawformer \cite{xiao2021lawformer} is used as the retriever. Lawformer, pre-trained based on Longformer \cite{beltagy2020longformer}, is specifically designed for understanding long Chinese legal documents. We use Lawformer to generate vector representations of queries and documents, and build an index of dense vectors. During retrieval, cosine distance is used to measure the similarity between queries and documents.

For the retrieval corpus, we use all statutes and judicial interpretations mentioned above from before and after the knowledge update as the retrieval corpus, prefixing statute names with "previous version of" or "latest version of" for differentiation. Each version of each statute is treated as a document to be retrieved, totaling 110,750 documents.

For the retrieval results, we use the top 3 most relevant documents, concatenating them as a prefix for the input prompt of LLMs.

\subsubsection{Parametric Update Strategies}

We implemented and tested two types of knowledge updating methods that involve the training of LLMs, namely model fine-tuning and model editing. 
% Since GLM4 is a close-source LLM, we cannot implement parametric update strategies on it.

\noindent \textbf{Model Fine-tuning}

We evaluate full fine-tuning (FT) and Lora fine-tuning \cite{hu2021lora}. We construct fine-tuning datasets for each model in the form of instructions, with the basic form as follows:

\begin{tcolorbox}[colback=lightgray!20,colframe=darkgray!80,title=Prompt for FT]
\textbf{User}: Please recite Article x of the Criminal Law of the People's Republic of China.

\textbf{Assistant}: Article x of the Criminal Law of the People's Republic of China: \{Content of the updated statute
\end{tcolorbox}

% \textit{User: Please recite Article x of the Criminal Law of the People's Republic of China.}

% \textit{Assistant: Article x of the Criminal Law of the People's Republic of China: \{Content of the updated statute\}}

During training, we only calculate the loss on the assistant's response part. We set the max token length to 1024. The detailed experimental settings for the two fine-tuning methods are as follows:

\begin{itemize}[leftmargin=*]
\item \textbf{Full Fine-tuning (FT)}: The learning rate is set to 1e-5, batch size to 16, and we use the AdamW optimizer\cite{loshchilov2019decoupled} for training over 4 epochs.
\item \textbf{Lora Fine-tuning}\cite{hu2021lora}: The learning rate is set to 1e-5, batch size to 16, and we use the AdamW optimizer for training over 6 epochs. For other hyperparameters in Lora fine-tuning, we set the decomposition order (r) to 32, the scaling parameter (Lora-alpha) to 64, and the dropout rate for Lora layers (Lora-dropout) to 0.05. The target module for Lora decomposition is the FFN layer of the Transformer model.
\end{itemize}

\noindent \textbf{Model Editing}

We evaluate three model editing methods: KN\cite{dai2021knowledge}, ROME\cite{meng2022locating}, and Self-Edit\cite{liu2024evedit}. The first two belong to triplet-level knowledge updates, while the last hopes to define editing scope through event reasoning anchor points to achieve event-level knowledge updates. For the first two knowledge editing methods, we use EasyEdit library \cite{wang2023easyedit} to edit the model.

\begin{itemize}[leftmargin=*]
\item \textbf{KN}\cite{dai2021knowledge}: The number of prompts for identifying knowledge neurons, n, is set to 10, the knowledge attribution threshold to 0.2, and the probability of retaining shared neurons to 0.4.
\item \textbf{ROME}\cite{meng2022locating}: We use the default settings of EasyEdit, uniformly setting the layer to be modified to the fifth layer and the layer for loss calculation to the last layer of the model. The weight decay rate is set to 1e-3.
\end{itemize}

When these two model editing methods are transferred to the legal domain, we define the format of the knowledge triplet as follows (the actual implementation is in Chinese):
\begin{equation}
\begin{aligned}
& (subject, relation, object) \\
& = (Statute Name, "Content\enspace is", Statute Content)
\end{aligned}
\nonumber
\end{equation}

\begin{itemize}[leftmargin=*]
\item \textbf{Self-Edit}\cite{liu2024evedit}: In this method, the original LLM is used to generate question-answer pairs related to updated knowledge, which are then used to fine-tune the model. We implement this method for legal knowledge updates and make necessary adjustments. Specifically, we ask the original LLM to pose a question about the updated statute, and then use nucleus sampling decoding with a probability threshold of 0.95 and a sampling temperature of 1.2 to sample 5 questions. For example:
\end{itemize}

\begin{tcolorbox}[colback=lightgray!20,colframe=darkgray!80,title=Prompt for question generation in Self-Edit]
\textbf{User}: The latest version of Article x of the Criminal Law of the People's Republic of China is: \{Content of the updated statute\}. Please pose a question about one aspect of this statute:

\textbf{Assistant}: The question is: ...
\end{tcolorbox}

% \textit{
% User: The latest version of Article x of the Criminal Law of the People's Republic of China is: \{Content of the updated statute\}. Please pose a question about one aspect of this statute:}

% \textit{Assistant: The question is: ...}

After obtaining the question, we use the original model to answer this question, forcing the prefix of the model's response to be the updated statute, and then have the model continue to write the answer. An example is as follows:

\begin{tcolorbox}[colback=lightgray!20,colframe=darkgray!80,title=Prompt for answer generation in Self-Edit]
\textbf{User}: Question: According to Article x of the Criminal Law of the People's Republic of China, \{Question Content\}?

\textbf{Assistant}: The new version of Article x of the Criminal Law of the People's Republic of China is: \{Content of the updated statute\}. Therefore, the answer to the question is:
\end{tcolorbox}

% \textit{
% User: Question: According to Article x of the Criminal Law of the People's Republic of China, \{Question Content\}?}

% \textit{Assistant: The new version of Article x of the Criminal Law of the People's Republic of China is: \{Content of the updated statute\}. Therefore, the answer to the question is:}

Then we use the QA pairs, along with the statutes, as the fine-tuning dataset. We only calculate the loss on the "Assistant" part in the above example.

Finally, we also test the models without using any knowledge update methods (abbreviated as \textbf{Raw} in the table), and with the context only including the corresponding statutes (abbreviated as \textbf{ICL-Golden} in the table), as baselines. All implementation details can be obtained from our public repository.\footnote{https://github.com/bebr2/LeKUBE}

\section{Experiment Results}

\subsection{Evaluation of Accuracy}

\begin{table*}[t!]
\centering
\setlength{\tabcolsep}{2.3pt}
\footnotesize{
\begin{tabular}{lcccc|cccc|cccc|ccccc}
\toprule
\textbf{Methods} & \multicolumn{4}{c}{\textbf{Recitation of Statutes(EM)}} & \multicolumn{4}{c}{\textbf{Recall of Statute Items(EM)}} & \multicolumn{4}{c}{\textbf{True-or-False Questions(Acc)}} & \multicolumn{4}{c}{\textbf{Multiple-Choice Questions(Acc)}} \\
\cmidrule{2-5} \cmidrule{6-9} \cmidrule{10-13} \cmidrule{14-17}
& \textbf{BC} & \textbf{CG} & \textbf{CL} & \textbf{LA} & \textbf{BC} & \textbf{CG} & \textbf{CL} & \textbf{LA} & \textbf{BC} & \textbf{CG} & \textbf{CL} & \textbf{LA} & \textbf{BC} & \textbf{CG} & \textbf{CL} & \textbf{LA} \\
\hline
\textbf{Raw} & 0.0111 & 0 & 0 & 0.0111 & 0.5056 & 0.0889 & 0.0389 & 0.4667 & 0.5530 & 0.6199 & 0.4953 & 0.6168 & 0.4222 & 0.4167 & 0.2722 & 0.3722 \\
\textbf{ICL-Golden} & 0.3222 & 0.3944 & \textbf{0.3444} & 0.5056 & \textbf{0.9889} & \textbf{0.9500} & \textbf{0.9333} & \textbf{0.7389} & \textbf{0.6682} & 0.5810 & \textbf{0.6838} & \textbf{0.6231} & \textbf{0.6444} & \textbf{0.4944} & \textbf{0.4500} & \textbf{0.4389} \\
\hline
\textbf{RAG-BM25} & 0.0722 & 0.1056 & 0.1222 & 0.1722 & 0.8722 & \uline{0.8500} & 0.8889 & 0.6667 & 0.5810 & 0.5763 & 0.5966 & 0.5794 & 0.4500 & 0.4056 & 0.2278 & 0.3500 \\
\textbf{RAG-LF} & 0.0111 & 0 & 0 & 0.0056 & \uline{0.9000} & \uline{0.8500} & \uline{0.9222} & \uline{0.6889} & 0.5109 & 0.5639 & 0.4938 & 0.5343 & 0.4333 & 0.3611 & 0.2944 & 0.3222 \\
\hline
\textbf{FT} & \uline{0.7667} & \uline{0.9278} & 0.0056 & \textbf{0.9389} & 0.4111 & 0.0722 & 0.0056 & 0.5167 & \uline{0.5841} & \uline{0.6277} & \uline{0.6090} & \uline{0.6215} & 0.4444 & 0.4000 & 0 & \uline{0.4000} \\
\textbf{Lora} & 0.0111 & 0 & 0 & 0.0167 & 0.4667 & 0.0889 & 0.0500 & 0.4667 & 0.5514 & 0.6199 & 0.4813 & 0.6184 & 0.4222 & 0.4111 & 0.2722 & 0.3500 \\
\hline
\textbf{Self-Edit} & \textbf{1.0000} & \textbf{0.9833} & \uline{0.2444} & \uline{0.7000} & 0.0056 & 0.0111 & 0 & 0.5222 & 0.5561 & 0.3146 & 0.5888 & 0.6153 & \uline{0.4667} & 0.3333 & \uline{0.3667} & 0.3778 \\
\textbf{KN} & 0.0111 & 0 & 0 & 0.0111 & 0.5056 & 0.0889 & 0.0389 & 0.4556 & 0.5498 & 0.6199 & 0.4969 & 0.6199 & 0.4222 & \uline{0.4167} & 0.2722 & 0.3667 \\
\textbf{ROME} & 0.0278 & 0 & 0.0167 & 0 & 0 & 0.0556 & 0 & 0 & 0 & \textbf{0.6355} & 0 & 0 & 0 & 0.3889 & 0 & 0 \\
\bottomrule
\end{tabular}
}
\caption{Evaluation of Accuracy and Generality: The first two tasks assess the accuracy and the last two tasks assess the generality. BC stands for BaiChuan2-13B-Chat, CG stands for ChatGLM3-6B, CL stands for ChatLaw-13B, LA stands for LegalAID-7B, RAG-LF stands for RAG-Lawformer, and FT stands for Full Fine-tuning. The best-performing knowledge update method under each model is in bold, while the second best is underlined.}
\label{tab:accandgen}
\vspace{-4mm}
\end{table*}

The evaluation of accuracy encompasses two tasks: the recitation of statutes and the recall of statute items. However, the performance of the same knowledge update method varies across these tasks.

\subsubsection{Recitation of Statutes}

Table \ref{tab:accandgen} presents the experimental results for evaluation of the accuracy and generality. From the first major column in Table \ref{tab:accandgen}, non-parametric strategies perform poorly on this task. According to statistics, the recall rate (i.e. the proportion of the corresponding statute included in the top 3 retrieved results) of Lawformer in this task is 0, but BM25, which is based on lexical matching, is 0.5167. Clearly, errors in the retrieval results of Lawformer mislead the model.
% Even models like GLM4 that support a 128k context still perform poorly.
The RAG-BM25 model also underperforms. Combined with the poor performance of ICL-Golden, it shows that it is difficult for the model to accurately recite statutes by updating in the context. 
The best performers in this task are full fine-tuning and the Self-Edit method in model editing, both of which change all the model parameters to fit the content of the statutes.
Lora fine-tuning, KN, and ROME models struggle to accurately remember the content of the statutes, with performance close to models without knowledge updates (i.e. Raw).

\subsubsection{Recall of Statute Items}

\begin{table}[!h]
    \centering
    \tiny
    \begin{tabular}{lll}
    \toprule
    \textbf{Knowledge Update Method}                     & \textbf{Model}            & \textbf{Proportion} \\ \hline
    \multirow{2}{*}{\textbf{Self-Edit}} & \textbf{Baichuan2Chat} & 0.0200            \\
                                        & \textbf{ChatGLM3}      & 0.2700            \\ \hline
    \multirow{3}{*}{\textbf{ROME}}      & \textbf{Baichuan2Chat} & 0                 \\
                                        & \textbf{ChatLaw}       & 0                 \\
                                        & \textbf{LegalAID}      & 0                 \\
    \bottomrule
    \end{tabular}
     \caption{Proportion of responses that the model follows instructions in the task of Recall of Statute Items. The value is calculated according to the proportion of model answers
containing statute item names (regardless of correctness).}
    \label{tab:recallinstruct}
\vspace{-3mm}
\end{table}

The second major column in Table \ref{tab:accandgen} presents the results for the recall of statute items. Interestingly, as the dual task of statute recitation, the performance of the knowledge update method in this task is inconsistent with that of statute recitation.
The best performance in this task is achieved by the baseline ICL-Golden, indicating that in this task, updating knowledge through additional context in the input prompts maybe enough. Since the retrieval methods of BM25 and Lawformer can easily retrieve the accurate statutes, leading to the excellent performance of the RAG method.
The five methods of parametric strategies do not show a performance that is significantly better than the models without knowledge updates, indicating that these methods might only fit the one-way connection from the statute item name to the statute content.
Furthermore, we delve into the reasons for the poor performance of Model Editing. We count the proportion of model answers containing statute item names (regardless of correctness), then show the proportion less than 0.98 in Table \ref{tab:recallinstruct}. We find that the Self-Edit and ROME methods severely impair the model's ability to follow instructions, leading to a decline in response accuracy.

\subsection{Evaluation of Generality}

\subsubsection{True-or-False Questions of Change in Statute}

The task assesses the capability of the updated LLMs to identify differences between historical and updated statutes.
From the third major column in Table \ref{tab:accandgen}, we can see that the best performance is achieved by ICL-Golden. In the parametric strategies, only the Full Fine-tuning (FT) shows superior performance to Raw. Other methods do not exhibit a clear improvement over the models which have not been updated. This suggests that the ability to compare differences in new and old knowledge remains a significant challenge for the commonly used parametric knowledge update methods.

\subsubsection{Multiple-Choice Questions of the Legal Scenario}

This task examines the reasoning and application of updated knowledge. 
As can be seen, from the last major column in Table \ref{tab:accandgen}, similar to the previous task, ICL-Golden performs the best. However, the low accuracy rates of all tested methods suggest that the task poses a significant challenge to some models' legal reasoning capabilities.
The performance of the two non-parametric strategies is not significantly better than the models that have not been updated. Among the parametric strategies, only Self-Edit performs better than the models that have not been updated. This suggests that simple fine-tuning and editing are insufficient for generalizing knowledge. It is difficult for models to acquire the reasoning capability of the knowledge with fitting of the knowledge. The slightly better performance of Self-Edit is because it includes the process of self-questioning and answering with new knowledge as the context in the training data, which clearly defines the reasoning process and the scope of reasoning influence. \cite{liu2024evedit}

\subsection{Evaluation of Locality}

\label{cha:local}

\begin{figure}[t]
    \centering
    \includegraphics[width=0.5\textwidth]{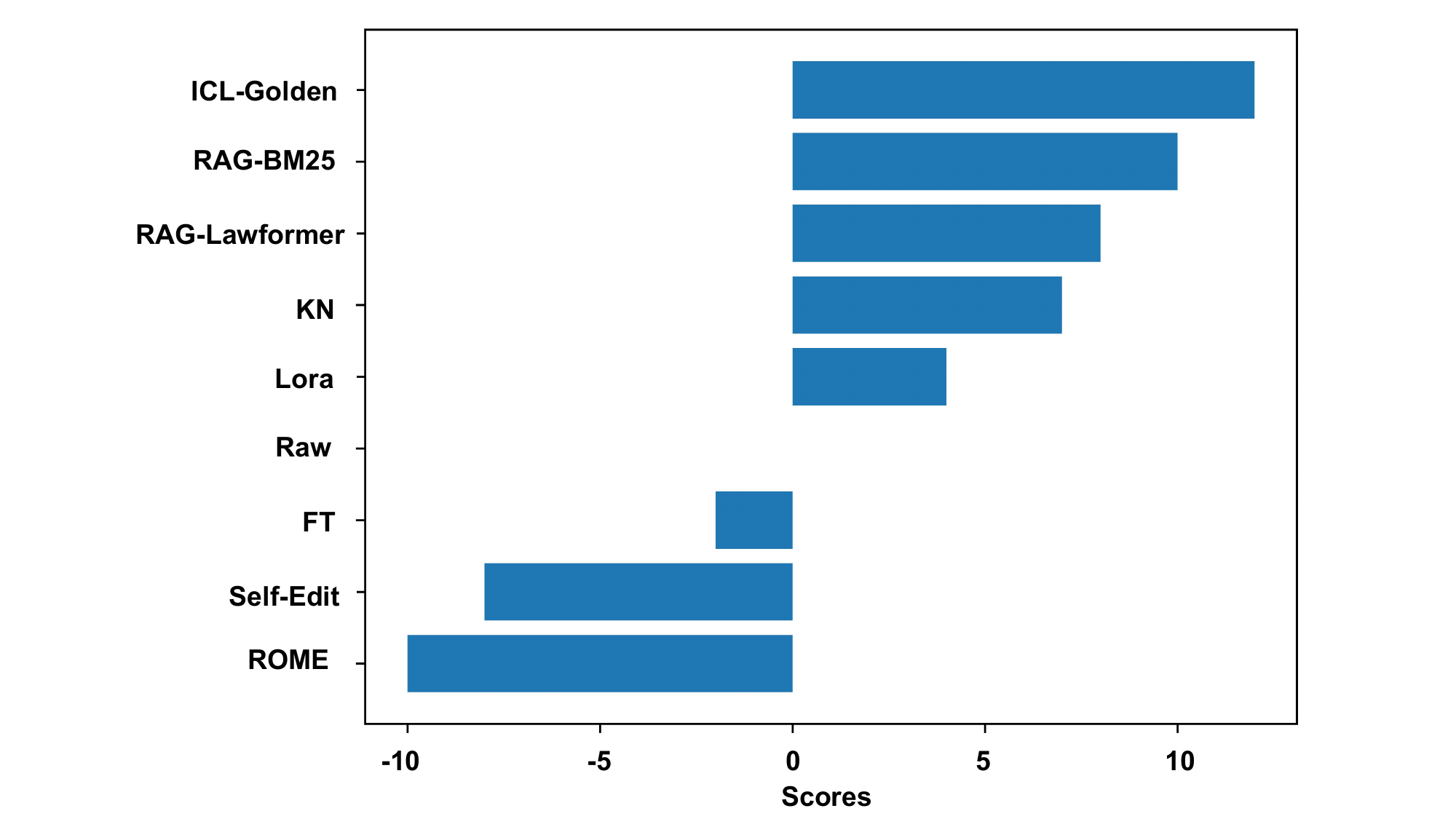}
      \caption{Evaluation of Locality. The calculated method of score is described in \autoref{cha:local}. The higher the score is, the better locality the method has.}
      \label{fig:local}
  \end{figure} 

Figure \ref{fig:local} summarizes the evaluation results of three tasks, namely recitation of statutes, recall of statute items, and true-or-false questions of change in statutes. It deserves to be noted that the statutes we used to test locality here are non-updated. The statistic on the horizontal axis is calculated as follows: we set the score of Raw as 0, and the initial scores for other methods are 0. For a certain task, if the performance of a certain knowledge update method applied on a certain LLM is not worse than that of the original LLM (i.e. Raw), the score is increased by one; otherwise, the score is decreased by one. Thus, the higher the score, the better the locality of the knowledge update method.

From Figure \ref{fig:local}, it can be intuitively seen that the locality of the non-parametric strategy is the best. One important reason is that the knowledge update of RAG only involves an update of the retrieval corpus and index, which has little impact on non-updated knowledge in both the retrieval corpus and the target LLM. Among the parametric strategies, the locality of the KN method is the best. Compared to full fine-tuning, Lora performs better since it only adds some low-rank structures to some layers of the LLM without modifying its original parameters. Taking ChatGLM3 as an example, in this experimental setting, the number of parameters trained only accounts for 0.1248\% of the total parameters of the model. This allows Lora to better maintain the model's memory of the original other knowledge. The locality of Self-Edit and ROME is very poor, indicating that the editing process has severely damaged the model's grasp of other knowledge.

\subsection{Evaluation of Scalability}

\begin{table*}[!t]
    \centering
    \resizebox{0.8\linewidth}{!} {
    \begin{tabular}{clllll}
    \toprule
    \textbf{}                        & \textbf{}    & \textbf{Baichuan2Chat}   & \textbf{ChatGLM3}     & \textbf{ChatLaw}         & \textbf{LegalAID}         \\ \hline
    \multirow{2}{*}{\textbf{Non-parametric}} & \textbf{RAG-BM25}    & 0.3889 / -0.0611         & 0.3333 / -0.0723      & 0.1944 / -0.0334         & 0.3333 / -0.0167          \\
                                     & \textbf{RAG-Lawformer} & 0.4333 / 0               & 0.3611 / 0            & 0.2944 / 0           & 0.3222 / 0                \\ \hline
    \multirow{2}{*}{\textbf{Model Fine-tuning}}   & \textbf{FT}             & 0.4222 / 0               & 0.4222 / \uline{0.0222} & 0.2722 / \textbf{0.2722} & 0.3667 / -0.0333      \\
                                     & \textbf{Lora}           & 0.4222 / 0               & 0.4000 / -0.0111         & 0.1500 / -0.1222           & 0.3667 / \uline{0.0167} \\ \hline
    \multirow{3}{*}{\textbf{Model Editing}}   & \textbf{Self-Edit}         & 0.4667 / 0               & 0.3778 / \uline{0.0445} & 0.3167 / -0.0500       & 0.3944 / \uline{0.0166} \\
                                     & \textbf{KN}             & 0.4222 / 0               & 0.4167 / 0        & 0.2722 / \textbf{0.2722} & 0.3667 / 0            \\
                                     & \textbf{ROME}           & 0.3556 / \textbf{0.3556} & 0.3944 / \uline{0.0055} & 0.0000 / 0                    & 0.3000 / \textbf{0.3000}        \\
    \bottomrule
    \end{tabular}
    }
\caption{Evaluation of Scalability: Multiple-Choice Questions of the Legal Scenarios. In each blank cell, the first value represents the accuracy after updating by chapters, and the second value is the difference between this accuracy and the accuracy after updating all related chapters. We bold the second value if it is greater than 0.05, and underline it if it is greater than 0 and less than or equal to 0.05. When the second value is greater than zero, the larger it is, the poorer scalability it is.}
\label{tab:scalmcq}
% \vspace{-4mm}
\end{table*}

% Table \ref{tab:scalmcq} presents the results of the scalability evaluation. In the table, we provide two values for each result. The first value represents the accuracy after updating by chapters, and the second value is the difference between the model's accuracy and the corresponding model in the last major column in Table \ref{tab:accandgen} (trained with knowledge updates in all corresponding chapters).
In this subsection, we compare two different experimental setups: the first involves processing each chapter by a separate LLM which only updates the knowledge of that particular chapter, with the final results being the average accuracy of all questions. The second setup involves processing all chapters by the same LLM which updates the knowledge of all chapters, and similarly calculates the average accuracy of all questions.\footnote{Figure \ref{fig:intro_scale} shows the process}. Table \ref{tab:scalmcq} shows the results, and the first value in each blank cell represents the result of the first setup, while the second one represents the difference between the first setup and the second setup. We focus on the second value. If this value is significantly greater than 0, it indicates that the knowledge update method has poor scalability, because its update performance on large datasets is significantly worse than that on small datasets.

The experimental results show that non-parametric strategies demonstrate good scalability, which is consistent with their good locality, as they only involve updates to the knowledge base. The performance of parametric strategies on large datasets declines to varying degrees compared to small datasets. Among them, the ROME method shows the poorest generality, which is most evident in the case of updating Baichuan2Chat and LegalAID. Lora fine-tuning demonstrates the best scalability in all parametric strategies.

\subsection{Evaluation of Retainability}

\label{cha:retain}

Table \ref{tab:retainmcq} presents the results of the retainability evaluation. In the table, $Acc_4$ represents the model accuracy after updating to the fourth subset (in the experimental setup, this point corresponds to the completion of the update of the criminal law), while $Acc_9$ represents the model accuracy after updating to the ninth subset (in the experimental setup, this point corresponds to the completion of the update of both the criminal and civil codes). The \textit{Diff.} column is the difference between $Acc_4$ and $Acc_9$. We focus on the \textit{Diff.} value. If this value is significantly greater than 0, it indicates that the knowledge updating method has poor retainability, and its performance after multiple updates is far worse than its performance after early updates.

\begin{table*}[t]
    \centering
    \resizebox{\linewidth}{!} {
    \begin{tabular}{clllllllllllll}
    \toprule
    &  & \multicolumn{3}{c}{\textbf{Baichuan2Chat}} & \multicolumn{3}{c}{\textbf{ChatGLM3}} & \multicolumn{3}{c}{\textbf{ChatLaw}} & \multicolumn{3}{c}{\textbf{LegalAID}} \\ 
    \cmidrule(lr){3-5} \cmidrule(lr){6-8} \cmidrule(lr){9-11} \cmidrule(lr){12-14}
    &  & \textbf{$Acc_4$} & \textbf{$Acc_9$} & \textit{\textbf{Diff.}} & \textbf{$Acc_4$} & \textbf{$Acc_9$} & \textit{\textbf{Diff.}} & \textbf{$Acc_4$} & \textbf{$Acc_9$} & \textit{\textbf{Diff.}} & \textbf{$Acc_4$} & \textbf{$Acc_9$} & \textit{\textbf{Diff.}} \\ \midrule
    \multirow{2}{*}{\textbf{Non-parametric}} & \textbf{RAG-BM25} & 0.4533 & 0.4533 & 0 & 0.3600 & 0.3600 & 0 & 0.2533 & 0.2533 & 0 & 0.3867 & 0.3867 & 0 \\
    & \textbf{RAG-Lawformer} & 0.4533 & 0.4533 & 0 & 0.2933 & 0.2933 & 0 & 0.3333 & 0.3333 & 0 & 0.3733 & 0.3733 & 0 \\ \midrule
    \multirow{2}{*}{\textbf{Model Fine-tuning}} & \textbf{FT} & 0.4933 & 0.5067 & -0.0134 & 0.3733 & 0.4133 & -0.0400 & 0.3067 & 0.3067 & 0 & 0.3867 & 0.4267 & 0 \\
    & \textbf{Lora} & 0.4800 & 0.4800 & 0 & 0.4133 & 0.4000 & \uline{0.0133} & 0.3200 & 0.3200 & 0 & 0.3867 & 0.3867 & 0 \\ \midrule
    \multirow{3}{*}{\textbf{Model Editing}} & \textbf{Self-Edit} & 0.4933 & 0.5467 & -0.0534 & 0.3333 & 0.2933 & \uline{0.0400} & 0.3733 & 0.4000 & -0.0267 & 0.4000 & 0.3867 & \uline{0.0133} \\
    & \textbf{KN} & 0.4800 & 0.4800 & 0 & 0.4133 & 0.4133 & 0 & 0.3067 & 0.3067 & 0 & 0.4000 & 0.4000 & 0 \\
    & \textbf{ROME} & 0.0400 & 0 & \uline{0.0400} & 0.3733 & 0.3600 & \uline{0.0133} & 0 & 0 & 0 & 0.2800 & 0 & \textbf{0.2800} \\ \bottomrule
    \end{tabular}
    }
    \caption{Evaluation of Retainability: Multiple-Choice Questions of the Legal Scenarios. We show  \textit{Acc4},  \textit{Acc9}, and \textit{Diff.} as evaluation metric, which is described in \autoref{cha:retain}. We bold the \textit{Diff.} values greater than 0.05 and underline those greater than 0 but less or equal to 0.05.  When the \textit{Diff.} value is greater than zero, the larger it is, the poorer retainability it is.}
    \label{tab:retainmcq}
    % \vspace{-4mm}
    \end{table*}

From the results, both non-parametric strategies and fine-tuning demonstrate good retainability, especially the non-parametric strategies. For model editing, compared to KN, ROME and Self-Edit show much worse performance in terms of retainability, which indicates that these types of knowledge edit methods couldn't edit LLM parameters precisely and new updates may overtake previous updates.

\subsection{Analysis of Time Cost}

\begin{table}[t]
    \centering
    \resizebox{0.85\linewidth}{!} {
        
    \begin{tabular}{clll}
    \toprule
    \multicolumn{1}{l}{}             &                        & \textbf{Train}       & \textbf{Inference} \\ \hline
    \textbf{Baseline}                      & \textbf{Raw}           & -                     & 124s            \\ \hline
    \multirow{2}{*}{\textbf{Non-parametric}} & \textbf{RAG-BM25}      & -                     & 909s            \\
                                     & \textbf{RAG-Lawformer} & -                     & 248s            \\ \hline
    \multirow{2}{*}{\textbf{Model Fine-tuning}}   & \textbf{FT}            & 1809s                 & 124s            \\
                                     & \textbf{Lora}          & 918s                  & 124s            \\ \hline
    \multirow{3}{*}{\textbf{Model Editing}}   & \textbf{Self-Edit}     & 10437s & 124s            \\
                                     & \textbf{KN}            & 2952s                 & 124s            \\
                                     & \textbf{ROME}          & 3400s                 & 124s            \\ \bottomrule
    \end{tabular}
    }
\caption{Comparison of time cost in for various knowledge update methods when updating Baichuan2-13B-Chat and conducting multiple-choice tasks.}
\label{tab:timestat}
\vspace{-6mm}
\end{table}

Table \ref{tab:timestat} provides a comparison of the time cost for various knowledge update methods when updating Baichuan2Chat and conducting multiple-choice tasks. For the non-parametric strategy, we did not include the time required to build the index; for the parametric strategy training process, the GPU resources used for model fine-tuning and Self-Edit are dual Nvidia A100(40G), while the GPU resources used for KN and ROME methods are quad Nvidia A100(40G). Inference was conducted on a single Nvidia A100(40G).

From the table, we can see that from the perspective of training, Self-Edit takes the longest time, mainly for two reasons:

\begin{itemize}[leftmargin=*]
  \item Self-Edit requires the model to first generate questions and answers through context learning, and multiple different questions need to be sampled, which already takes 1603s;
  \item During training, the data involved in Self-Edit is much more than other parametric methods, which will take longer time under the same training epoch, consuming 8843s.
\end{itemize}

In addition, the other two model editing methods are not fast either, taking much longer than the time for model fine-tuning, which contradicts the claim of "efficient editing" of model editing. We speculate that the efficiency of model editing is reflected in a single or a small number of updates, and for a large range of knowledge updates (e.g., LeKUBE's update involves 180 statutes), model editing does not have an advantage. Moreover, many frameworks for model fine-tuning reduce resource consumption or accelerate training, such as Deepspeed\cite{deepspeed}, while the tools for model editing are still mainly considering integrating various editing methods and ensuring that editing is applicable to more models.

In model fine-tuning, the advantages of Lora fine-tuning in low resource consumption and fast speed are obvious in our experiments. Even if the number of training epochs is twice that of full-parameter fine-tuning, the time consumed is only half of that of full-parameter fine-tuning.

\section{Conclusions and Future Works}

In this paper, we conduct an in-depth analysis of knowledge update methods applied in legal LLMs and propose a novel legal knowledge update benchmark, LeKUBE, to evaluate the performance of these methods.
% We have identified the unique challenges posed by the legal domain, including the complexity and specificity of legal knowledge, which make the application of general domain knowledge update methods to the legal sphere particularly challenging. In response to this, we have developed LeKUBE, a benchmark that covers key aspects of legal knowledge updating and thoroughly evaluates the five fundamental properties of knowledge update techniques: accuracy, generality, locality, scalability, and retainability. Our experiments have confirmed that LeKUBE presents a significant challenge to existing knowledge update methods and effectively differentiates their capabilities in legal knowledge updating.
% Our study also suggests several directions for future research.
% \begin{itemize}
%   \item As for the benchmark, LeKUBE only evaluates the most basic capability of "comparing new and old knowledge differences" for the legal applicability issue. Future tasks could consider more complex challenges related to legal applicability, which would require the participation of more legal experts.
%   \item As for knowledge update techniques, existing parametric strategies could be improved to better meet the demands of the legal domain. The exploration of a combination of parametric and non-parametric strategies might lead to the development of update techniques that are more suitable for legal knowledge updating.
% \end{itemize}
Our study highlights potential areas for further investigation. Firstly, LeKUBE only evaluates the most basic capability of "comparing new and old knowledge differences" for the legal applicability issue. Future tasks could consider more complex challenges related to legal applicability, which would require the participation of more legal experts.
Secondly, existing parametric strategies could be improved to better meet the demands of the legal domain. The exploration of a combination of parametric and non-parametric strategies might lead to the development of update techniques that are more suitable for legal knowledge updating.

% \clearpage
\bibliographystyle{unsrt}
\bibliography{sample-base}

\end{document}